\documentclass[letterpaper, 10 pt, conference]{ieeeconf}
\usepackage{times}
\usepackage[pdftex]{graphicx}
\usepackage{amsmath,amssymb,amsopn,amstext,amsfonts}
\usepackage{cancel}
\usepackage[space]{cite}
\usepackage{pdfsync}
\usepackage{balance}
\usepackage{color}
\usepackage{mathtools}
\usepackage{algpseudocode}
\usepackage{algorithm} 
\usepackage{bm}

\usepackage{diagbox}
\usepackage{float}
\usepackage{epstopdf}
\usepackage{pifont}
\usepackage{fixltx2e}
\usepackage{amsmath}
\usepackage{multirow}
\usepackage{url}
\usepackage{verbatim}
\usepackage[linkcolor=black,citecolor=black,urlcolor=black,colorlinks=true]{hyperref}
\usepackage{booktabs}
\usepackage{graphicx}
\usepackage{subcaption}

\algnewcommand\algorithmicforeach{\textbf{for each}}
\algdef{S}[FOR]{ForEach}[1]{\algorithmicforeach\ #1\ \algorithmicdo}
\newcommand{\tp}{^{\mathsf{T}}}

\graphicspath{{figures/}}
\DeclareGraphicsExtensions{.png,.jpg,.eps,.pdf}
\IEEEoverridecommandlockouts
\title{\LARGE \bf STD-Trees: Spatio-temporal Deformable Trees for Multirotors Kinodynamic Planning}
\author
{Hongkai Ye, Chao Xu, and Fei Gao
	\thanks{
		All authors are with State Key Laboratory of Industrial Control Technology, Zhejiang University, Hangzhou 310027, China 
		and with Huzhou Institute of Zhejiang University, Huzhou 313000, China.}
	\thanks{Email: \tt(hkye, cxu, and fgaoaa)@zju.edu.cn}
	\thanks{
		Code: \url{https://github.com/ZJU-FAST-Lab/std-trees}}
}

\begin{document}
	\maketitle
	\thispagestyle{empty}
	\pagestyle{empty}
	
	\begin{abstract}
		In constrained solution spaces with a huge number of homotopy classes, stand-alone sampling-based kinodynamic planners suffer low efficiency in convergence. Local optimization is integrated to alleviate this problem. 
		
		In this paper, we propose to thrive the trajectory tree growing by optimizing the tree in the forms of deformation units, and each unit contains one tree node and all the edges connecting it.
		The deformation proceeds both spatially and temporally by optimizing the node state and edge time durations efficiently.
		The unit only changes the tree locally yet improves the overall quality of a corresponding sub-tree. Further, variants to deform different tree parts considering the computation burden and optimizing level are studied and compared, all showing much faster convergence. 
		The proposed deformation is compatible with different RRT-based kinodynamic planning methods, and numerical experiments show that integrating the spatio-temporal deformation greatly accelerates the convergence and outperforms the spatial-only deformation.
		
	\end{abstract}
	
	\section{Introduction}
	\label{sec:introduction}
	Kinodynamic planning\cite{Donald1993kinodynamic} respects directly the system kinematics and the dynamical constraints in the process of globally searching for an optimal trajectory in exploring the entire constrained solution space and bring an optimal trajectory. 
	Although much progress has been made, it remains challenging to globally plan a near-optimal trajectory efficiently in large-scale environments with complex obstacles.
	
	For multirotor systems of high dimensional states, sampling-based variants\cite{Schmerling2015drift, webb2013kinodynamic, Hongkai2021tgk} are popular by holding the anytime property and being asymptotically optimal. 
	They connect state samples in the discretized state space and grow trajectory trees to explore the constrained solution space.
	By introducing some randomness, they are naturally suitable to search into spaces near different local minima and potentially bring better solutions as more computation resources are available. 
	Although the randomness necessities optimality, it impedes the convergence since the probability to sample exactly near the optimal solution is extremely low, especially in high dimensional solution spaces where a huge number of homotopy classes exist. 
	
\begin{figure}[t]
		\centering
		\begin{subfigure}{0.45\linewidth}
			\includegraphics[width=1\linewidth]{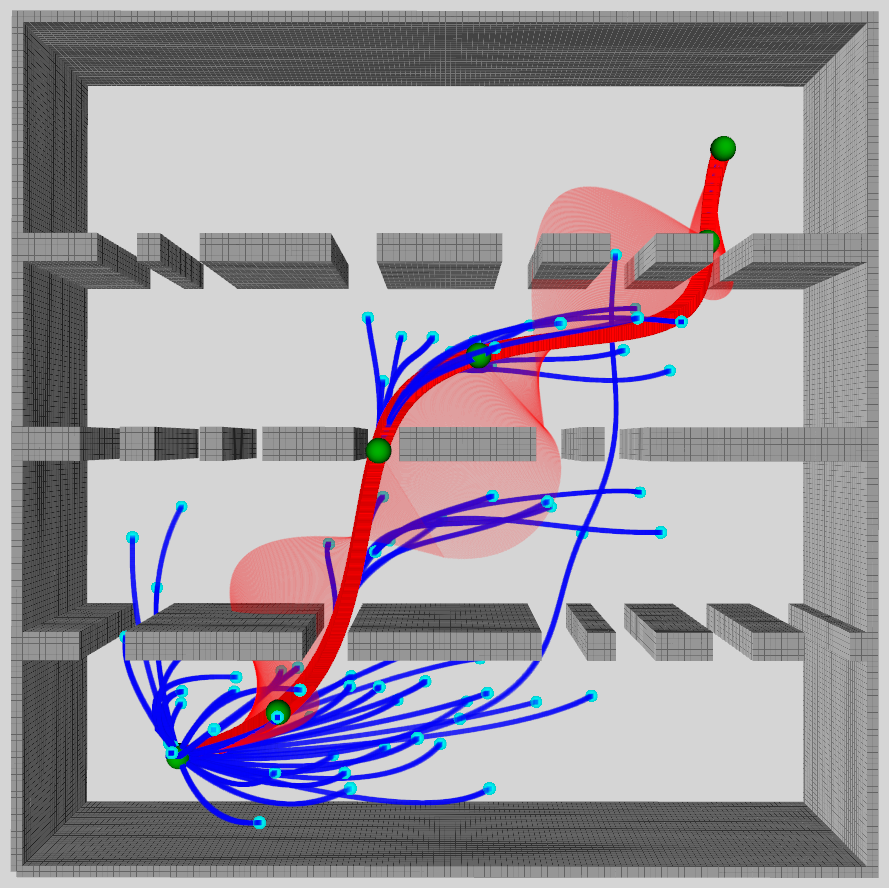}	
			\caption{With ST Deformation.}
			\label{subfig:regional_ap}
		\end{subfigure}
		\begin{subfigure}{0.45\linewidth}
			\includegraphics[width=1\linewidth]{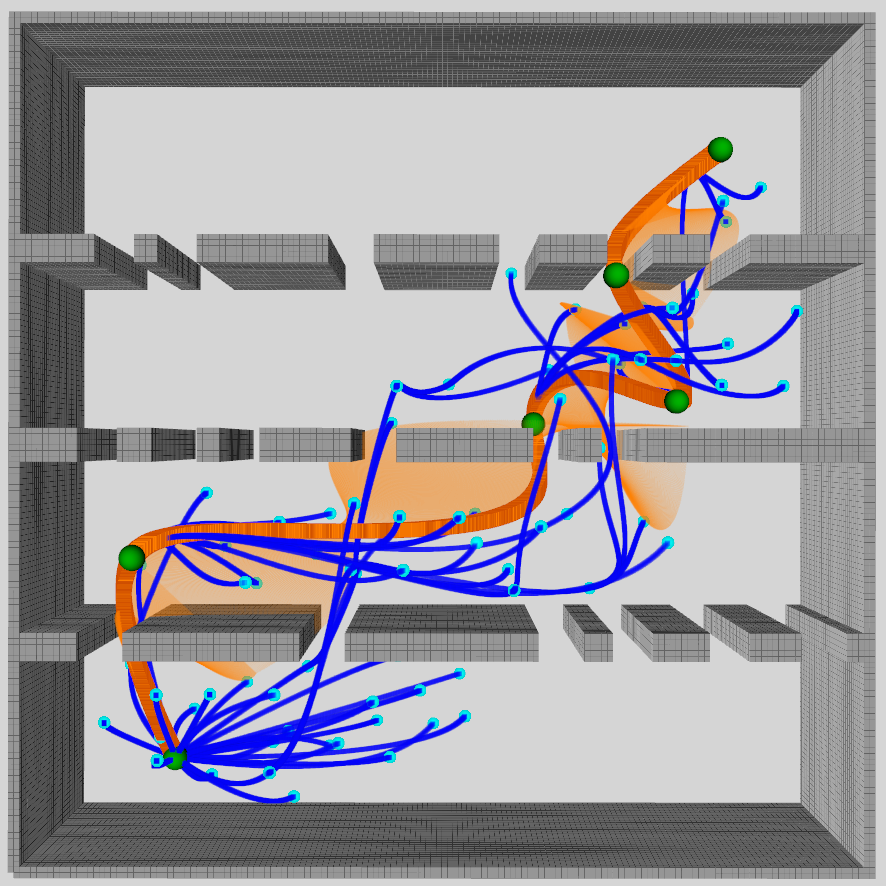}
			\caption{Without deformation.}
			\label{subfig:backend_ap}
		\end{subfigure}
		\captionsetup{font={small}}
		\caption{Current tree and best trajectory after adding a same number of nodes. Planning with deformation acquires a better-organized tree structure with overall higher quality and gets a smoother trajectory.}
		\label{fig:tree_compare}
		\vspace{-0.77cm}
	\end{figure}

	We embrace local optimization to alleviate this problem.
	The optimization is performed by locally deform the trajectory tree and keep it well grown, both spatially and temporally, as shown in Fig.~\ref{fig:tree_compare}.
	Tree deforming is studied by Hauer and Tsiotras~\cite{Hauer-RSS-17} in the geometric path planning problem. 
	They reposition the locations of nodes to minimize the overall tree cost and improve the convergence.
	However, they do not account for trajectory planning.
	Focused on kinodynamic planning problems where an extra time dimension needs to be considered, we show that deforming the trajectory tree in additional the time dimension, that is, spatio-temporally (ST), can further facilitate the convergence. 
	The difficulties, however, appear in designing practical objectives and attaining efficiency. Therefore, we propose to deform the tree in forms of deformation units to balance the optimization level and the computation burden.
	
	In this work, we focus on accelerating the convergence in sampling-based kinodynamic planning by optimizing specific node states and the time durations of some related edges.
	We summarize the main contributions as follows:
	\begin{enumerate}
		\item We propose a novel sampling-based kinodynamic planner for multirotors.
		It thrives the trajectory tree growing by optimizing the tree in forms of deformation unit that contains one tree node and the edges connecting to it. One deformation unit only constitutes a small part of the entire tree, yet deforming it improves the overall quality of the sub-tree rooted at the deformed node. 
		\item We propose to efficiently refine the deformation unit in not only shapes but also the time dimension. It is compatible with different RRT-based kinodynamic planners, and integrating it shows a faster convergence and generates much better final trajectories.
		\item We propose several variants to perform the deformation by different levels of computation burden. Numerical results show that all the variants converge quicker than the baseline and overtake some search-based planners.
	\end{enumerate}

	\section{Related Work}
	
	Basically, different sampling-based algorithms differ in two procedures, the exploration of solution space and the exploitation of information brought by new samples, resulting in different convergence.
	The original RRT algorithm~\cite{LaValle2001randomized} simply connects a state node to the nearest node in the current tree and is very fast in exploration to get solutions. 
	It is, however, proved to converge to a sub-optimal solution by Karaman and Frazzoli in ~\cite{karaman2011} where the RGG and RRT* algorithms are introduced. 
	They stand out in the process of choosing parent nodes and rewire local tree structure according to the best \textit{cost-from-start} values and can converge to the optimal with probability one. 
	The RRT$^\#$~\cite{Arslan2013rrtsharp} algorithm resolves the under-exploitation in RRT* and improves convergence by further rewiring in cascade to propagate the new information to other parts of the tree and maintain a well-grown tree of consistent nodes. 
	We adopt the RRT$^\#$ scheme to balance the exploration-exploitation trade-off in our kinodynamic algorithm framework. 
	To acquire higher convergence rate, recent works focus on improving sampling strategies by sampling directly in the Informed Set~\cite{informedRRTStar2014iros}, the Relevant Region Set~\cite{Sagar2021Relevant}, or the Local Subset of GuILD~\cite{Aditya2021Guided}. 
	However, these designs depend on the $L2$-Norm objective. For applications with varied desired objectives, analytic forms of the sampling set can not be obtained. 
	
	Combining local optimization in the global searching process is studied by many for geometric shortest path planning. 
	RABIT*~\cite{Choudhury2016Regionally} adopts CHOMP~\cite{chomp2013ijrr} to
	find feasible alternative connections for edges in collision and thus accelerate the search. However, the convergence rate highly depends on shrinking the $L2$-Norm based informed set in their BIT*~\cite{bitStar2015icra}-like batch sampling and searching process. The rather heavy optimization can also hinder exploring different homotopy classes.
	DRRT~\cite{Hauer-RSS-17} does not require any informed sampling set.
	It minimizes the overall tree cost by optimizing the locations of some existed nodes in the tree. A well-grown tree is maintained to cover the search space in each iteration, and thus paths obtained are of high quality and the convergence is improved. 
	For kinodynamic planning, INSAT~\cite{Natarajan2021Interleaving} combines trajectory optimization into discretized graph search. However, the solving time is intractable because of high-dimensional graph search, and long trajectories are being optimized many times in each node expanding step.
	Although weighted A*~\cite{Pohl1970First, Ebendta2009Weighted} is used to
	accelerate searching, it eventually generates a sub-optimal solution since the admissibility criterion is relaxed.
	In our combining scheme, the sampling-based search globally reasons for the optimal solution with asymptotic optimality, and the optimization only deforms limited edges of the tree. 
	
	\begin{algorithm}[t]
		\caption{Spatio-temporal Deformable Trees}
		\label{alg:kino_rrt_star}
		\begin{algorithmic}[1]
			\State \textbf{Notation}: Tree $\mathcal{T}$,\ State $\mathbf{x}$,\ Deformation Units $\mathcal{U}$,\ Environment $\mathcal{E}$,\ Deform Type $\mathcal{L} \in$ \{NODE, TRUNK, BRANCH, TREE\}
			\State \textbf{Initialize}: $\mathcal{T} \leftarrow \emptyset \cup \{\mathbf{x}_{start}\}$
			\While{Termination condition not met}
			\State $\mathbf{x}_{new} \leftarrow$ \textbf{Sampling}($\mathcal{E}$)
			\State $\mathcal{X}_{backward} \leftarrow$ \textbf{BackwardNear}($\mathcal{T}$, $\mathbf{x}_{new}$)
			\State $\mathbf{x}_n \leftarrow$ \textbf{ChooseParent}($\mathcal{X}_{backward}$, $\mathbf{x}_{new}$)
			\State $\mathcal{T} \leftarrow \mathcal{T} \cup \{\mathbf{x}_n, \mathbf{x}_{new}\}$
			\If{\textbf{TryConnectGoal}($\mathbf{x}_{new}$, $\mathbf{x}_{goal}$)}
			\State One Solution Found.
			\EndIf
			\State $\mathcal{U} \leftarrow $ \textbf{SelectDeformationUnits}($\mathbf{x}_n$, $\mathcal{L}$)
			\State \textbf{DeformInOrder}($\mathcal{U}$)
			\State \textbf{RewireInCascade}($\mathcal{T}$, $\mathbf{x}_{new}$)
			\EndWhile
			\State \Return $\mathcal{T}$
		\end{algorithmic}
	\end{algorithm}

	\section{Problem Statement}
	\label{problem_statement}
	The multirotors kinodynamic planning problem is to find a trajectory that minimizes some cost and satisfies some constraints. 
	Following\cite{webb2013kinodynamic, Hongkai2021tgk, liu2017iros, boyu2019ral}, the cost minimized is a trade-off between time and energy, which suits many applications. The constraints imposed include obstacle avoidance, system dynamics, kinematics and start and goal constraints.
	According to the multirotor systems' differential flatness property\cite{MelKum1105}, we can use a linear model of chain integrator to represent its dynamics with four flat outputs $p_x, p_y, p_z$ (position in each axis), $\psi$ (yaw), and their derivatives being the state variables.
	Thus, the trajectory planning problem is formulated as a Linear Quadratic Minimum Time (LQMT) problem in the area of optimal control. For each axis (yaw is excluded in this paper), it is formulated as follows:
	\begin{equation}
		\label{equ:problem_form}
		\begin{split}
			\min_{u(t)} \mathcal{J}&=\int_{0}^{\tau}(\rho+\frac{1}{2}u(t)^2)dt \\[1ex]
			s.t. \quad & \mathbf{A}\mathbf{x}(t)+\mathbf{B}u(t) - \mathbf{\dot{x}}(t)=\mathbf{0}, \\
			&\mathbf{x}(0)=\mathbf{x}_{start}, \
			\mathbf{x}(\tau)=\mathbf{x}_{goal}, \\
			&\mathcal{G}(\mathbf{x}(t), u(t)) \preceq \mathbf{0}, \ \forall t \in [0, \tau],
		\end{split}
	\end{equation}
	where $\tau$ is the time duration of the trajectory, $\rho$ the trade-off weight to penalize time against energy, 
	$\mathbf{x}_{start}$ the initial state, $\mathbf{x}_{goal}$ the goal state, 
	and $\mathcal{G} \preceq \mathbf{0}$ the obstacle avoidance and higher derivative limit constraints. 
	
	\section{kinodynamic Deformable Trees}
	\label{std-tree}
	The nonlinear constraints make the entire solution space highly non-convex. 
	Directly applying nonlinear programming will most definitely fall into local minima.
	We adopt a sampling-based approach as described in Alg.~\ref{alg:kino_rrt_star} to address the problem. 
	Basically, the algorithm is based on kinodynamic RRT*~\cite{webb2013kinodynamic} with features of RRT$^\#$~\cite{Arslan2013rrtsharp} and with an extra step to deform the shape and the time duration of the current trajectory tree (line 12) in each iteration.
	The searching framework breaks the problem down into smaller subproblems by incrementally adding state samples to probe the solution space. 
	Each subproblem is of the same form as Equ.\ref{equ:problem_form} with different boundary states, and we get a trajectory segment by solving one.
	
	In each iteration, after a state node is sampled, we seek its best parent node in the BackwardNear node set $\mathcal{X}_{backward}$ according to a \textit{cost-from-start} value estimated by solving the LQMT subproblems. 
	If a qualified best parent is found, the sampled node and a new edge are added to the tree. It then tries to connect to the goal node for a possible solution. After that, the core part, the tree deformation is performed by optimizing some trajectory edges, both the profile and the time duration. This step reduces the overall tree cost without adding more nodes. 
	Since spending more time on exploitation means less exploration, the deformation is only activated after finding a first feasible solution. 
	Finally, since the structure of the implicit graph is changed while adding nodes, we rewire some nodes for potential improvement of their \textit{cost-from-start} values. The rewire performs in cascade to propagate the new information brought by the new node to other parts of the entire tree.
	
	The final trajectory tree consists of many trajectory segments as its edges, and the optimal trajectory between start and goal consists of a chain of trajectory segments between a series of successive states.
	
	\section{Tree Edge Representation}
	As seen, solving the LQMT subproblems makes up a fundamental part of the searching process and will be called many times in each iteration. Therefore, it needs to be solved fast. 
	Suppose the integrator model is of $s^{th}$-order, which leads the state variable to be $\mathbf{x}(t)=[p(t), \cdots, p^{(s-1)}(t)]\tp$ and the control to be $u(t)=p^{(s)}(t)$. 
	According to~\cite{Lewis1991LQMT, webb2013kinodynamic}, if all the inequality constraints are temporally disregarded, the optimal control law for the unconstrained LQMT problems are parameterized by polynomials of degree $s-1$, and then each state variable in $\mathbf{x}(t)$ can be obtained by integration up to $s$ times.
	Thus, we have $p(t) = \mathbf{c}\tp\beta(t), t \in [0,T]$, where $\mathbf{c} \in \mathbb{R}^{2s}$ is the polynomial coefficient vector, $T$ the time duration, and $\beta(t)=(1,t,t^2,\cdots, t^{2s-1})\tp$ the natural basis.
	
	This leads to a conclusion that given time durations $T$ and boundary conditions $\mathbf{d}=[\mathbf{x}\tp(t)|_{t=0}, \mathbf{x}\tp(t)|_{t=T}]\tp \in \mathbb{R}^{2s}$, the optimal solutions for these specific LQMT problems are fully determined since no freedom is left.
	As claimed by our previous work~\cite{wang2020generating}, a smooth bijection exists between the polynomial coefficient vector $\mathbf{c}$ and the boundary condition vector $\mathbf{d}$, that is, an analytic transformation exist between the two kinds of descriptions of polynomials \{$\mathbf{c}, T$\} and \{$\mathbf{d}, T$\}:
	\begin{equation}
		\mathbf{d} = \mathbf{A}_f(T)\mathbf{c},\ \mathbf{c}=\mathbf{A}_b(T)\mathbf{d},
	\end{equation}
	where $\mathbf{A}_f(T)$ and $\mathbf{A}_b(T)$ are forward and backward mapping matrices whose entries are analytically determined.
	The cost of one edge can then be calculated by \{$\mathbf{c}, T$\} as
	\begin{equation}
		\begin{split}
			\label{equ:edge_cost_c_T}
			J_s(\mathbf{c}, T) 
			&= \rho T + \int_0^T{\frac{1}{2} \mathbf{c}\tp \beta^{(s)}(t) \beta^{(s)}(t)\tp \mathbf{c}}\ dt \\
			&= \rho T + \frac{1}{2}\mathbf{c}\tp \mathbf{Q}(T) \mathbf{c},
		\end{split}
	\end{equation}
	or by \{$\mathbf{d}, T$\}  as
	\begin{equation}
		\begin{split}
			\label{equ:edge_cost_d_T}
			J_s(\mathbf{d}, T) =&\ J_s(\mathbf{x}(t)|_{t=0},\ \mathbf{x}(t)|_{t=T},\ T) \\
			=&\ \rho T + \frac{1}{2} \mathbf{d}\tp \mathbf{M}(T) \mathbf{d}, \\
			\mathbf{d} = 
			\begin{bmatrix}
				\mathbf{x}(t)|_{t=0} \\
				\mathbf{x}(t)|_{t=T} 
			\end{bmatrix},& \ 
			\mathbf{M}(T) = \mathbf{A}\tp_b(T) \mathbf{Q}(T) \mathbf{A}_b(T),
		\end{split}
	\end{equation}
	both analytically.
	The previously ignored constraints are checked afterwards. If satisfied, a tree edge is grown, and the cost provides an estimation of the optimal transition cost between two state samples in the constrained solution space.
	
	The two kinds of representation of trajectory edges support the following tree deformation process.

	\section{Trajectory Tree Deformation}
	\label{sec:method}
	
	\begin{figure}[t]
		\centering
		\vspace{0.25cm}
		\includegraphics[width=0.8\linewidth]{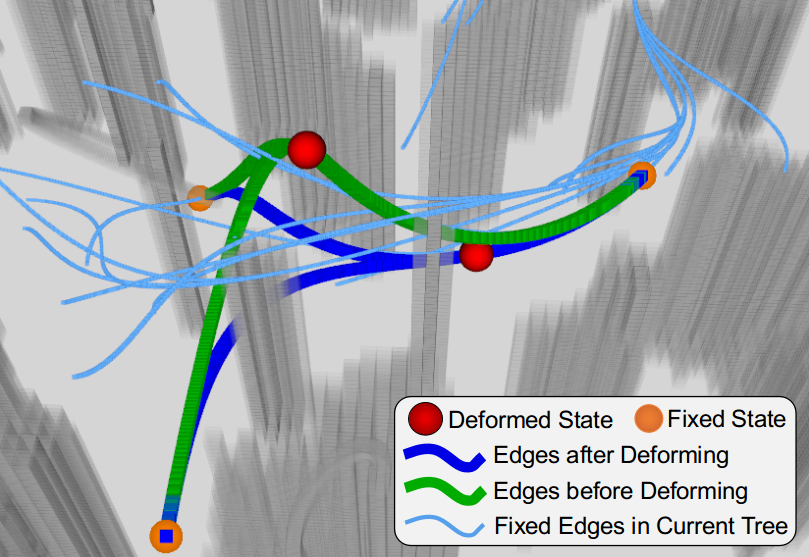}
		\captionsetup{font={small}}
		\caption{The deforming of one deformation unit. In this case, the deformed node has two child nodes, which are fixed together with the parent node. Only the time durations of the deformed edges and the state of the deformed node are optimized.}
		\label{deform_process}
		\vspace{-0.5cm}
	\end{figure}
	
	By the end of each iteration of the tree growing process, some new information of the solution space is collected in the tree. However, the \textit{cost-from-start} value of some nodes may not well estimate the actual value due to insufficient sampling, presenting that the tree does not grow well, especially in the time dimension. A deforming of the edges can improve its quality, as depicted in Fig.~\ref{deform_process}.
	
	\subsection{Deformation Unit}
	Given current tree structure $\mathcal{T}$, each node $n \in \mathcal{T}$ contains the following information:
	\begin{itemize}
		\item $\mathbf{x}_n$: the corresponding state of $n$,
		\item $p_n$: the parent node of $n$ in $\mathcal{T}$, 
		\item $T_n$: the time duration of the edge from $p_n$ to $n$, 
		\item $\mathbf{c}_n$: the coefficient vector of the edge from $p_n$ to $n$, 
		\item $\mathcal{C}_n$: the children node set of $n$ in $\mathcal{T}$, 
		\item $g_n$: the \textit{cost-from-start} value of $n$ following $\mathcal{T}$. 
	\end{itemize}
	
	If we reposition one of the nodes and fix all others, meanwhile keeping the connection between nodes unchanged, then only the edges that connect to the repositioned node will be affected. 
	Thus, we set a deformation unit by the collection of one node, the edge from its parent, and the edges to its children.
	Within a unit, the deformation is performed by optimizing the node state and the time duration of these edges, while all other nodes' state and edge time durations are fixed.
	We denote the element collection in a deformation unit as 
	\begin{equation}
		\{\mathbf{x}_n, \mathbf{T}_n\} = \{\mathbf{x}_n, T_n, T_i,| i \in \mathcal{C}_n\}, 
	\end{equation}
	where $n$ is the node being deformed, and $\mathbf{T}_n = \{T_n, T_i,| i \in \mathcal{C}_n\}$ contains the time durations of all the edges connecting to $n$.
	By setting the node state as decision variables instead of edge coefficients, the equality constraints imposed by the continuity requirement in-between tree nodes are implicitly eliminated, and the number of decision variables is reduced. 
	
	\subsection{Objective Design}
	When optimizing the deformation unit, we need an appropriate objective. 
	It should follow that after the deformation, the overall tree quality is improved and it stands a better chance to find a better solution or the optimal solution.
	As revealed by Hauer et al.~\cite{Hauer-RSS-17}, the sum of the \textit{cost-from-start} values of a fixed number of sample nodes stands for an estimation of the optimal value function for the solution space covered by the selected sample set.
	Considering also that optimizing the deformation unit will and only will change the \textit{cost-from-start} values of node $n$ and all its descendant nodes, leaving all other nodes unaffected.
	We therefore set the objective as 
	\begin{equation}
		g_n + \sum_{i \in \mathcal{D}_n}{g_i}, 
	\end{equation}
	where $\mathcal{D}_n$ is the set of all $n$'s descendant nodes. 
	Optionally, we can assign a weight $w \in [0,1]$ for each node suggesting the possibility of it constituting the final best trajectory. A heuristic estimation of the \textit{cost-to-go} can be an alternative for this weight.
	The objective suggests an optimization of the sub-tree rooted at node $n$.
	
	By growing of the tree, we have 
	\begin{equation}
		g_n = c(\mathbf{x}_{p_n}, \mathbf{x}_{n}) + g_{p_n},
	\end{equation}
	where $c(\mathbf{x}_i, \mathbf{x}_j)$ is the edge cost connecting node $i$ to node $j$, analytically estimated by Equ.~\ref{equ:edge_cost_d_T}.
	We can then use the weighted sum of all the edge costs in the tree to calculate the objective.
	The weight of an edge $(i, j)$ is assigned by counting the total number of paths that start from the start node to any other node in the sub-tree using this edge. 
	The objective thus becomes 
	\begin{equation}
		\begin{split}
			g_n + \sum_{i \in \mathcal{D}_n}{g_i} &= \sum_{i \in \mathcal{T}_n} \sum_{j \in \mathcal{C}_i}d_j c(\mathbf{x}_i, \mathbf{x}_j) \\ 
			&= d_n c(\mathbf{x}_{p_n}, \mathbf{x}_n) + \sum_{i \in \mathcal{C}_n} d_i c(\mathbf{x}_n, \mathbf{x}_i) + \mathbf{C} \\
			&= \sum_{i \in \{n\}\bigcup \mathcal{C}_n} d_i c(\mathbf{x}_{p_i}, \mathbf{x}_i) + \mathbf{C}, 
		\end{split}
	\end{equation}
	where $\mathcal{T}_n$ is the sub-tree rooted at node $n$, $d_n$ equals $1 + nb\_des(n)$ with $nb\_des(n)$ the number of descendants of node $n$, and $\mathbf{C}$ a constant indicating the sum of all other edges' cost in the tree, which is irrelevant to the elements in the deformation unit.
	Therefore, the objective depends only on the edges in the deformation unit yet expresses the quality of a sub-tree that estimates the value function of a part of the solution space.
	
	\subsection{Unconstrained Formulation}
	In the process of tree growing, we ignore the constraints and check afterwards to achieve faster exploration of the entire solution space. 
	When deforming, it is wished to focus more on the local space covered by the deformation unit and exploit the information already gathered in the tree.
	The constraints on obstacle avoidance and dynamical feasibility should now be considered.
	
	Denote $\mathcal{G}(p^{[s]}(t)) \in \mathbb{R}^{s+1} \preceq \mathbf{0}$ the functional-type constraints that consider obstacle avoidance and dynamical limitations such as the velocity, acceleration and other higher order derivative constraints up to order $s$. 
	To eliminate these inequality constraints, we construct penalty functions and turn them into soft ones. 
	Since they consist of an infinite number of inequality constraints and can hardly be directly handled, we transform them into finite-dimensional ones via integral of constraint violations, and the integral is estimated by weighted sum of the sampled penalty functions.
	Note that in practice, these constraints are decoupled between edges, that is, $\mathcal{G}(p^{[s]}(t))$ with $t \in [0, T_i]$ are solely determined by $\mathbf{c}_i$ and $T_i$.
	Thus, for one trajectory edge parameterized by \{$\mathbf{c}_i, T_i$\}, we compute the penalty function as
	\begin{equation}
		J_f(\mathbf{c}_i, T_i, k_i)=\frac{T_i}{k_i} \sum_{j=0}^{k_i}\omega_j \mathcal{X}\tp max[\mathcal{G}(\mathbf{c}_i, T_i, t),\ \mathbf{0}], 
	\end{equation}
	where $k_i$ is the sample number on this edge, $\mathcal{X} \in \mathbb{R}_{\geq 0}^{s+1}$ is a vector of penalty weights for each entry of $\mathcal{G}$, $(\omega_0, \omega_1, \cdots, \omega_{k_i - 1}, \omega_{k_i}) = (1/2, 1, \cdots, 1, 1/2)$ are the quadrature coefficients following the trapezoidal rule, $max[\ ]$ is an element-wise comparison, and $t = j/k_i \cdot T_i$ is the sampled time stamp.
	Integrating the penalty into the objective by each edge and the unconstrained optimization of one deformation unit is formulated as
	\begin{equation}
		\begin{split}
			\min_{\mathbf{x}_n, \mathbf{T}_n} &\sum_{i \in \{n\}\bigcup \mathcal{C}_n} d_i (J_s(\mathbf{c}_i, T_i) + J_f(\mathbf{c}_i, T_i, k_i)), \\
			&\mathbf{c}_i = 
			\left\{
			\begin{array}{cl}
				\mathbf{A}_b(T_i)
				\begin{bmatrix}
					\mathbf{x}\tp_{p_n}, \ 
					\mathbf{x}\tp_n 
				\end{bmatrix}\tp, & i = n \\
				\mathbf{A}_b(T_i)
				\begin{bmatrix}
					\mathbf{x}\tp_n, \
					\mathbf{x}\tp_i 
				\end{bmatrix}\tp, & i \in \mathcal{C}_n
			\end{array}
			\right.
		\end{split}
	\end{equation}
	with \{$\mathbf{x}_n, \mathbf{T}_n$\} being the decision variables. 
	
	\subsection{Spatio-temporal Optimization}
	For one edge \{$\mathbf{c}_i, T_i$\}, $i \in \{n\}\bigcup \mathcal{C}_n$ in the deformation unit, we derive the gradient of the decoupled objective w.r.t \{$\mathbf{x}_n, \mathbf{T}_n$\} by chain rule as follows: 
	\begin{flalign}
		&\frac{\partial J_s}{\partial \mathbf{x}_n} = \frac{\partial J_s}{\partial \mathbf{c}_i} \frac{\partial \mathbf{c}_i}{\partial \mathbf{x}_n} = \mathbf{Q}(T)\mathbf{c}_i, 
		\frac{\partial J_f}{\partial \mathbf{x}_n} = \frac{\partial J_f}{\partial \mathcal{G}} \frac{\partial \mathcal{G}}{\partial \mathbf{c}_i} \frac{\partial \mathbf{c}_i}{\partial \mathbf{x}_n},&
	\end{flalign}
	\begin{flalign}
		&\frac{\partial J_s}{\partial T_i} = \rho + \frac{1}{2} \mathbf{c}_i\tp \dot{\mathbf{Q}}(T_i) \mathbf{c}_i, 
		\frac{\partial J_f}{\partial T_i} = \frac{J_f}{T_i} + \frac{\partial J_f}{\partial \mathcal{G}} \frac{\partial \mathcal{G}}{\partial t} \frac{j}{k_i},&
	\end{flalign}
	\begin{flalign}&
		\frac{\partial J_f}{\partial \mathcal{G}} = \frac{T_i}{k_i}\sum_{j=0}^{k_i}\omega_j \mathcal{X} \odot max[Sign[ \mathcal{G}(\mathbf{c}_i, T_i, \frac{j}{k_i})],\ \mathbf{0}], &
	\end{flalign}
	\begin{flalign}&
		\frac{\partial \mathbf{c}_i}{\partial \mathbf{x}_n} = 
		\left\{
		\begin{array}{cl}
			\begin{bmatrix}
				\mathbf{A}_b^{01}(T_i)\tp \
				\mathbf{A}_b^{11}(T_i)\tp
			\end{bmatrix}\tp, & i = n \\
			\begin{bmatrix}
				\mathbf{A}_b^{00}(T_i)\tp \
				\mathbf{A}_b^{10}(T_i)\tp
			\end{bmatrix}\tp, & i \in \mathcal{C}_n, 
		\end{array}
		\right.&
	\end{flalign}
	where $\odot$ is the Hadamard product, $Sign[\ ]$ is element-wise indication of the sign of a number, and $\mathbf{A}_b^{xx}(T)$ contains block elements of $\mathbf{A}_b(T)$ partitioned by $s$. 
	The gradients of an edge w.r.t. the time durations of other edges are all $0$.
	The conditional computation implies different cases that $\mathbf{x}_n$ being tail or head state of an edge.
	
	The only parts undefined in the above equations are $\partial \mathcal{G}/\partial \mathbf{c}_i$ and $\partial \mathcal{G}/\partial t$. 
	For obstacle avoidance, we wish the edge positions have certain clearance and thus define 
	\begin{equation}
		\mathcal{G}_o = r - \mathcal{F}(p_i(t)),
	\end{equation}
	where $r>0$ is the preferred distance away from obstacles, and $\mathcal{F}(p(t))$ computes a minimum distance to the closest obstacle given a position in the edge, which can be pre-computed with a distance field built incrementally~\cite{Oleynikova2017Voxblox, Han2019FIESTA} or in batch~\cite{Felzenszwalb2012Distance}.
	For dynamical feasibility, we limit the amplitude of higher-order derivatives and define 
	\begin{equation}
		\mathcal{G}_k = p_i^{(k)}(t)^2 - m_k^2,\quad k \in \{1, 2, \cdots, s\}
	\end{equation}
	where $m_k$ is the maximum allowed value of the $k^{th}$ derivative.
	The corresponding gradients are 
	\begin{equation}
		\begin{split}
			&\frac{\partial \mathcal{G}_k}{\partial \mathbf{c}_i} = 2\beta^{(k)}(t)p_i^{(k)}(t)\tp, \quad \frac{\partial \mathcal{G}_k}{\partial t} = 2\beta^{(k+1)}(t)\tp \mathbf{c}_i p_i^{(k)}(t), \\
			&\frac{\partial \mathcal{G}_o}{\partial \mathbf{c}_i} = -\beta(t) \frac{\partial \mathcal{F}}{\partial p_i(t)}, \quad 
			\frac{\partial \mathcal{G}_o}{\partial t} = -\dot{\beta}(t)\tp \mathbf{c}_i \frac{\partial \mathcal{F}}{\partial p_i(t)}, \\
		\end{split}
	\end{equation}
	where $\partial \mathcal{F} / \partial p(t)$ is computed with interpolation of the distance field. 
	The analytic transformation between \{$\mathbf{c}_i,T_i$\} and \{$\mathbf{x}_{p_i}, \mathbf{x}_i, T_i$\} provides efficiency for calculating the objective and the gradients since no matrix inversion is required. 
	With the gradients at hand, numerical optimization is applied then to solve this optimization problem. 
	Considering that the interpolation of the distance fields $\mathcal{F}$ introduces non-smoothness, Newton-type methods like BFGS can be inaccurate or fail sometimes. We thus adopt bundle methods to address it. 
	In this work, the Limited Memory Bundle Method (LMBM)~\cite{Haarala2007GloballyCL} is used.
	
	\subsection{Deformation Variants}
	\label{variants}
\begin{figure}[t]
		\centering
		\vspace{0.3cm}
		\includegraphics[width=1\linewidth]{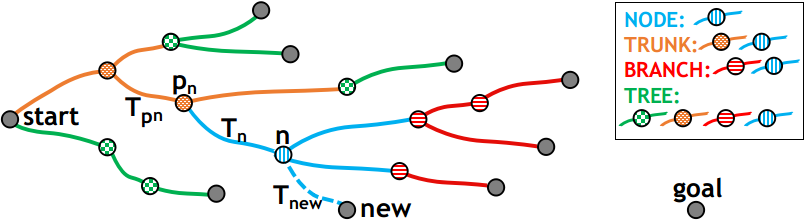}
		\captionsetup{font={small}}
		\caption{Deformation variants depiction. The start node and all the leaf nodes stay fixed.}
		\label{deform_type}
		\vspace{-2.4cm}
	\end{figure}

	The optimization of one deformation unit can explore its surrounding local solution space more thoroughly while other parts of the tree are left unchanged. 
	After each iteration of Alg.~\ref{alg:kino_rrt_star} though, the current tree topology in different parts of the tree can be changed by adding new nodes and by \textit{RewireInCascade}.
	Applying deformation to more tree parts can improve the overall tree quality but may at the cost of more computation loads. 
	Weighing the optimization levels and the computation burden, we propose four kinds of variants.
	As depicted in Fig.~\ref{deform_type}, after one node $new$ is newly added to the tree, denoting its parent node in the tree by $n$, the tree deformation is performed as one of the following variants.
	\begin{enumerate}
		\item \textbf{NODE}: optimizes only one deformation unit which contains $n$ and the edges connecting to it.
		\item \textbf{TRUNK}: optimizes several deformation units, and the units are selected by following parent pointers from $n$ up to a child of $start$. The deformation is then performed in an order from the child node to $n$. This is the strategy DRRT~\cite{Hauer-RSS-17} adopts.
		\item \textbf{BRANCH}: optimizes every nodes and edges in the sub-tree rooted at $n$. The order follows breadth-first search starting from $n$. All the leaf nodes are excluded.
		\item \textbf{TREE}: optimizes every node and edge in the entire tree except the start node and all the leaf nodes. The order follows breadth-first search starting from the children nodes of $start$.
	\end{enumerate}
	
	\section{Numerical Results}
	\label{sec:experiment}
	\subsection{Experiment Settings}
	
	\begin{figure*}[t]
		\centering
		\includegraphics[width=0.999\linewidth]{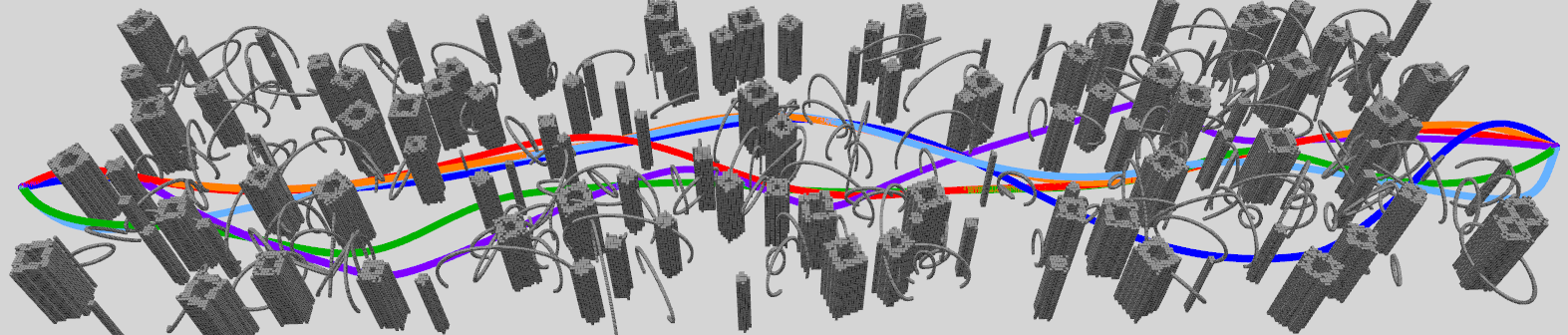}
		\captionsetup{font={small}}
		\caption{Environment for the variants comparison and final trajectories of one trial. The color indications are light blue for NODE, orange for TRUNK, red for BRANCH, green for TREE, purple for w/o, and dark blue for search-based with a weight $1.7$. The proposed variants generate smoother trajectories with lower costs, especially variant BRANCH (red).}
		\label{fig:forest_env_variants}
	\end{figure*}
	
	\begin{figure*}[t]
		\centering
		\includegraphics[width=0.999\linewidth]{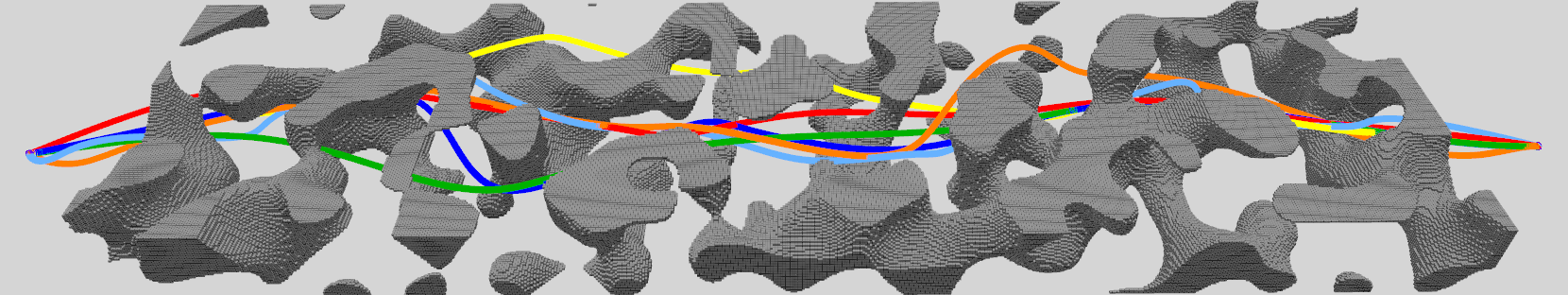}
		\captionsetup{font={small}}
		\caption{Environment for the experiment in Sec.~\ref{subsec:exp2} and final trajectories of one trial. The color indications are light blue for kRRT, dark blue for kRRT-ST, yellow for kRRT*, green for kRRT*-ST, orange for kRRT$^\#$, and red for kRRT$^\#$-ST. Planning with ST deformation generates smoother trajectories with lower cost (red, green, and dark blue compared with orange, yellow, and light blue, respectively).}
		\label{fig:cave_env_variants}
	\end{figure*}
	
	\begin{figure}[t]
		\centering
		\includegraphics[width=0.93\linewidth]{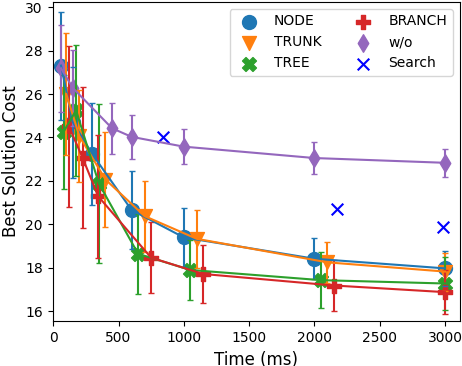}
		\captionsetup{font={small}}
		\caption{Convergence comparisons of deformation variants and search-based methods. Short lines indicate standard deviations. No heuristic amplification costs tens of seconds to search (not shown).}
		\label{fig:convergence_variants}
	\end{figure}
	
	\begin{figure}[t]
		\centering
		\includegraphics[width=0.96\linewidth]{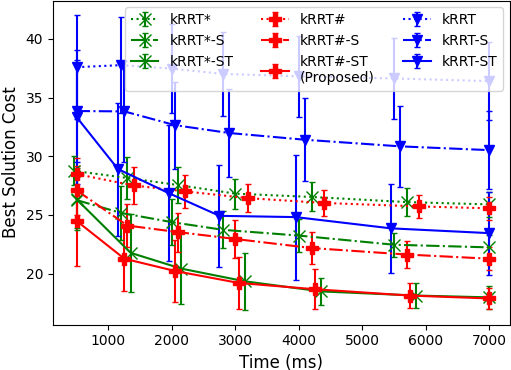}
		\captionsetup{font={small}}
		\caption{Convergence of different searching schemes with the proposed spatio-temporal deformation (-ST) and with the spatial-only deforming (-S) compared with stand-alone sampling-based kinodynamic planners. Short lines indicate standard deviations.}
		\label{fig:convergence_wo}
		\vspace{-0.5cm}
	\end{figure}
	
	For numerical comparisons, we set $s=3$, which means a third-order integrator is used to model our multirotor system. The weight of time $\rho$ is set $100$. 
	The dynamical limitations are set as $5m/s$ for velocity, $7m/s^2$ for acceleration, and $15m/s^3$ for jerk. 
	For collision checking, the maps are pre-built and transformed into occupancy grids of $0.1m$ resolution with obstacles inflated by $0.2m$.
	All the numerical experiments are conducted with a desktop computer with a 3.4GHz Intel i7-6700 processor.
	
	\subsection{Variants Comparison}
	In this experiment, the proposed planning method w/o deformation or with one of the four deformation variants introduced in Sec.~\ref{variants} are benchmarked with each other and with search-based kinodynamic planners. 
	For search-based ones, the weighted version of A*~\cite{Ebendta2009Weighted} is used to accelerate the search. Three different weights are used to amplify the heuristic: $1.7, 2.3,$ and $2.8$.
	Each variant runs for $100$ trials with a $3$ second time budget. The environment and final trajectories of one trial are shown in Fig.~\ref{fig:forest_env_variants}. 
	As Fig.~\ref{fig:convergence_variants} shows, the proposed methods find the first solution within milliseconds and then quickly converge.
	Compared to the baseline (w/o), the convergence rate all improves remarkably by planning with any variants of the proposed spatio-temporal deformation.
	Among the variants, BRANCH slightly beats TREE and much outperforms NODE and TRUNK. We think it is because a new sample brings potential improvements mostly on the sub-tree while other tree parts are less likely influenced, and thus deforming just the sub-tree is a good balance on computation cost and tree quality improvement.
	The search-based ones with different weights, however, take much longer time to get a solution with higher cost.
	
	\subsection{Deformation Comparison}
	\label{subsec:exp2}
	For three sampling-based kinodynamic planning methods, kRRT, kRRT*, and kRRT$^\#$, we compare the proposed spatio-temporal deformation with deforming only spatially and with no deformation, denoting a suffix of -ST, -S, and no suffix, respectively. 
	For each deformation, variant BRANCH is adopted.
	Each method runs for $100$ trials with a $7$ second time budget. The environment and final trajectories of one trial are shown in Fig.~\ref{fig:cave_env_variants}. 
	Fig.~\ref{fig:convergence_wo} shows that for all the RRT-based methods, both deformations accelerate the convergence evidently. 
	The proposed spatio-temporal deformation further considerably outperforms the spatial-only deforming .

	\section{Conclusion}
	\label{sec:discussion}
	In this paper, we propose a sampling-based kinodynamic planning method for multirotors combining local optimization by deforming some selected tree edges spatially and temporally. The optimization is performed efficiently within deformation units which contain the state of one node and the time durations of the edges connecting it. Though only limited parts are deformed, the overall quality of a sub-tree is improved and a well-grown tree is maintained after each node is added to the tree. By deforming the state of some selected nodes and related edge durations without adding more nodes, the convergence is improved. Benchmark results show that integrating the proposed deformation achieves a much faster convergence rate.
	We open source our code for the reference of the community.
	
	\bibliography{ICRA2022_Hongkai}
\end{document}